\documentclass[conference]{IEEEtran}
\IEEEoverridecommandlockouts
\usepackage{cite}
\usepackage{amsmath,amssymb,amsfonts}
\usepackage{algorithmic}
\usepackage{graphicx}
\usepackage{textcomp}
\usepackage{xcolor}
\usepackage{anyfontsize}
\usepackage{algorithm,algorithmic}
\def\BibTeX{{\rm B\kern-.05em{\sc i\kern-.025em b}\kern-.08em
    T\kern-.1667em\lower.7ex\hbox{E}\kern-.125emX}}

\begin{document}

\title{Sensor Fusion for Autonomous Indoor UAV Navigation in Confined Spaces\\

}

\author{\IEEEauthorblockN{Alice James\textsuperscript{1}\IEEEauthorrefmark{1}, Avishkar Seth\textsuperscript{1}, Endrowednes Kuantama\textsuperscript{2}, Subhas Mukhopadhyay\textsuperscript{1}, Richard Han\textsuperscript{2}}
\IEEEauthorblockA{\textit{School of Engineering\textsuperscript{1}, School of Computing\textsuperscript{2}} \\
\textit{Macquarie University}\\
Sydney, Australia \\
alice.james@mq.edu.au}
\thanks{Article Published: doi: 10.1109/ICST59744.2023.10460820}
}

\maketitle

\begin{abstract}

In this paper, we address the challenge of navigating through unknown indoor environments using autonomous aerial robots within confined spaces. The core of our system involves the integration of key sensor technologies, including depth sensing from the ZED 2i camera, IMU data, and LiDAR measurements, facilitated by the Robot Operating System (ROS) and RTAB-Map. Through custom designed experiments, we demonstrate the robustness and effectiveness of this approach. Our results showcase a promising navigation accuracy, with errors as low as 0.4 meters, and mapping quality characterized by a Root Mean Square Error (RMSE) of just 0.13 m. Notably, this performance is achieved while maintaining energy efficiency and balanced resource allocation, addressing a crucial concern in UAV applications. Flight tests further underscore the precision of our system in maintaining desired flight orientations, with a remarkable error rate of only 0.1\%. This work represents a significant stride in the development of autonomous indoor UAV navigation systems, with potential applications in search and rescue, facility inspection, and environmental monitoring within GPS-denied indoor environments. 

\end{abstract}


\section{Introduction}
In recent years, the rapid advancements in Unmanned Aerial Vehicle (UAV) technology have ushered in a new era of exploration and data acquisition, particularly in complex indoor environments where traditional navigation methods often fall short. These aerial robots, particularly quadrotors equipped with a multitude of sensors, have the potential to revolutionize tasks such as search and rescue \cite{Bi2019ARescue}, facility inspections \cite{wawrla2019applications}, and environmental monitoring within indoor spaces due to their enhanced mobility and maneuverability \cite{Tao2023LearningVehicles}. The difficulty of autonomous navigation in constrained, GPS-denied indoor environments is substantial \cite{Elmokadem2021TowardsSurvey, Labbe2022Jun}. The combination of Deep Learning (DL) and multi-sensor fusion is crucial in improving the navigation \cite{WangYangWangLiuXu+2013+466+481}, mapping, and exploring capacities of UAVs \cite{QianMaDaiFang+2012+295+314}. 

\par The problem of autonomous indoor robot mapping, localization, and navigation is widely studied in the literature. Simultaneous Localization and Mapping, or SLAM, is a fundamental concept in this domain that relies heavily on robot sensing and perception. This process is particularly challenging indoors due to the absence of GPS signals, the presence of various obstacles, and the need for precise, real-time decision-making \cite{Li2022High-AccuracyEnvironments}. UAVs operating in indoor spaces encounter tight corridors, dynamic environments, and limited space to maneuver, all of which intensify the complexity of navigation. Additionally, UAVs designed for indoor use face constraints in terms of onboard computational power and payload capacity \cite{Wilson2021Dec, Sandino2020Oct}. Despite these challenges, the demand for indoor applications, such as surveillance and disaster relief in GPS-denied environments \cite{Boonyathanmig}, is on the rise. In such scenarios, the preference is often for low-cost flying robots that can be easily replaced if damaged or lost \cite{Tao2023LearningVehicles}.

\par Real-time Simultaneous Localisation and Mapping (RTAB-Map) is an open-source package that focuses on loop closure detection and effective memory management for large-scale mapping \cite{Labbe2018, Labbe2014Sep, Labbe2013Feb}. It provides online processing, significant low-drift odometry, robust localisation, and multi session mapping. These features were developed to address various practical requirements. 

\par In this paper, a deep learning-enhanced multi-sensor fusion SLAM system for autonomous indoor environment exploration of UAVs is proposed. This paper focuses on the development of the UAV based on multiple sensors such as IMU, vision and laser sensors and fusing the data with deep learning based computer vision sensing. This paper's main contributions are as follows:

\begin{itemize}
\item   \textbf{Unique UAV system}: A novel UAV system architecture is proposed that uses customised open source software and hardware configurations with the on-board computer, flight controller and sensors.

\item   \textbf{Multi Sensor-Fusion}: A new approach is proposed that creates a sensor fusion using the LIDAR, camera, and IMU sensing modalities.

\item \textbf{Deep Learning}: A deep learning architecture for autonomous indoor navigation in tight spaces is presented that uses visual slam to create edge detection.


\end{itemize}

\section{Background}

Previous research on autonomous indoor navigation for UAVs using deep learning and SLAM techniques is very significant in ground based robots \cite{Kolhatkar2020Jul, Filipenko}. The objective of these works is to address the unique challenges and limitations inherent to UAV navigation within indoor settings

\par Additionally, systems that merge Light Detection and Ranging (LIDAR) with Inertial Measurement Units (IMU) have been suggested for indoor UAV navigation \cite{Kumar2017Jun}. Utilizing a pair of scanning laser range finders along with an IMU, this method offers robust navigation by allowing the UAV to precisely gauge its environment. Moreover, advancements in multi-sensor fusion and depth learning for UAV navigation have also been examined \cite{Liu2022May}. Exploiting the improvements in sensor technology, this method provides real-time data support to deep learning algorithms, thereby making autonomous drone navigation feasible.

\par A great deal of research has been done on 2D laser range finders \cite{Zhang2020May}, 3D LIDAR \cite{BibEntry2023Sep}, and vision sensors \cite{Sumikura2019Oct, MacarioBarros2022Feb} over the past few years. Recent studies have focused on a variety of different tools that can be used for SLAM, including magnetic \cite{Viset2022Apr}, olfactory \cite{Chen2023Jun}, and thermal sensors \cite{Saputra2021Nov}, and event-based cameras \cite{Chamorro2022Jun}. However, these substitute sensors have not yet been taken into account for SLAM in the same extent as range and vision sensors.

\par In conclusion, the existing frameworks for deep learning-based autonomous indoor navigation for UAVs include a wide range of approaches. These comprise deep reinforcement learning, obstacle avoidance, integration of LIDAR and IMU, multi-sensor fusion, and comparative evaluations. Each aims to surmount the obstacles and limitations related to indoor UAV navigation, thereby facilitating secure and efficient autonomous operation. While the existing area of research has significantly advanced the field of autonomous indoor navigation for UAVs, our work introduces an integrated approach that suggests an alternative to current solutions.

\section{System Overview}

\begin{figure}[ht]    
    \centering
    \includegraphics[width=1\linewidth]{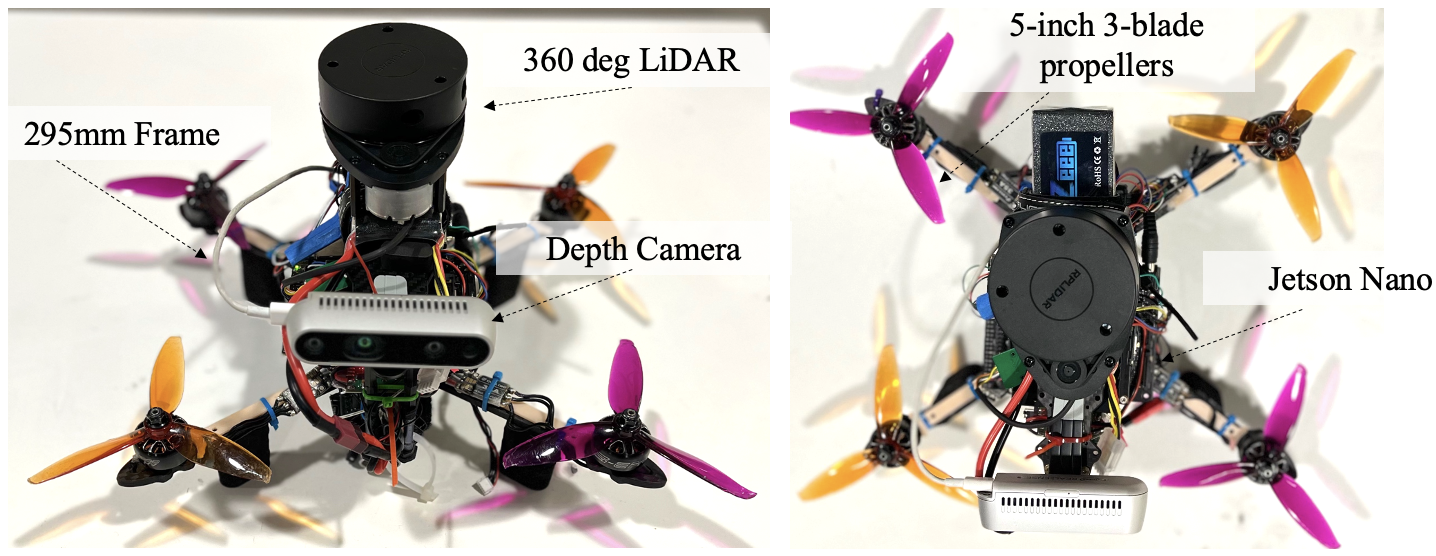}
	\caption{The aerial robot used for indoor SLAM}
\label{drone}  
\end{figure}

In this study, the challenge of autonomous aerial robot's state estimation is tackled as a hardware software integrated issue. The objective of this system is to enable a quadcopter to autonomously navigate through indoor environments, essentially replicating the decision-making capabilities of a skilled human pilot. A front-facing stereo camera mounted on the UAV generates images that are processed in real-time. Using a trained classifier, the system issues flight commands aimed at ensuring safe navigation shown in Figure~\ref{drone}. 

\par
\begin{figure}[ht]    
    \centering
    \includegraphics[width=1\linewidth]{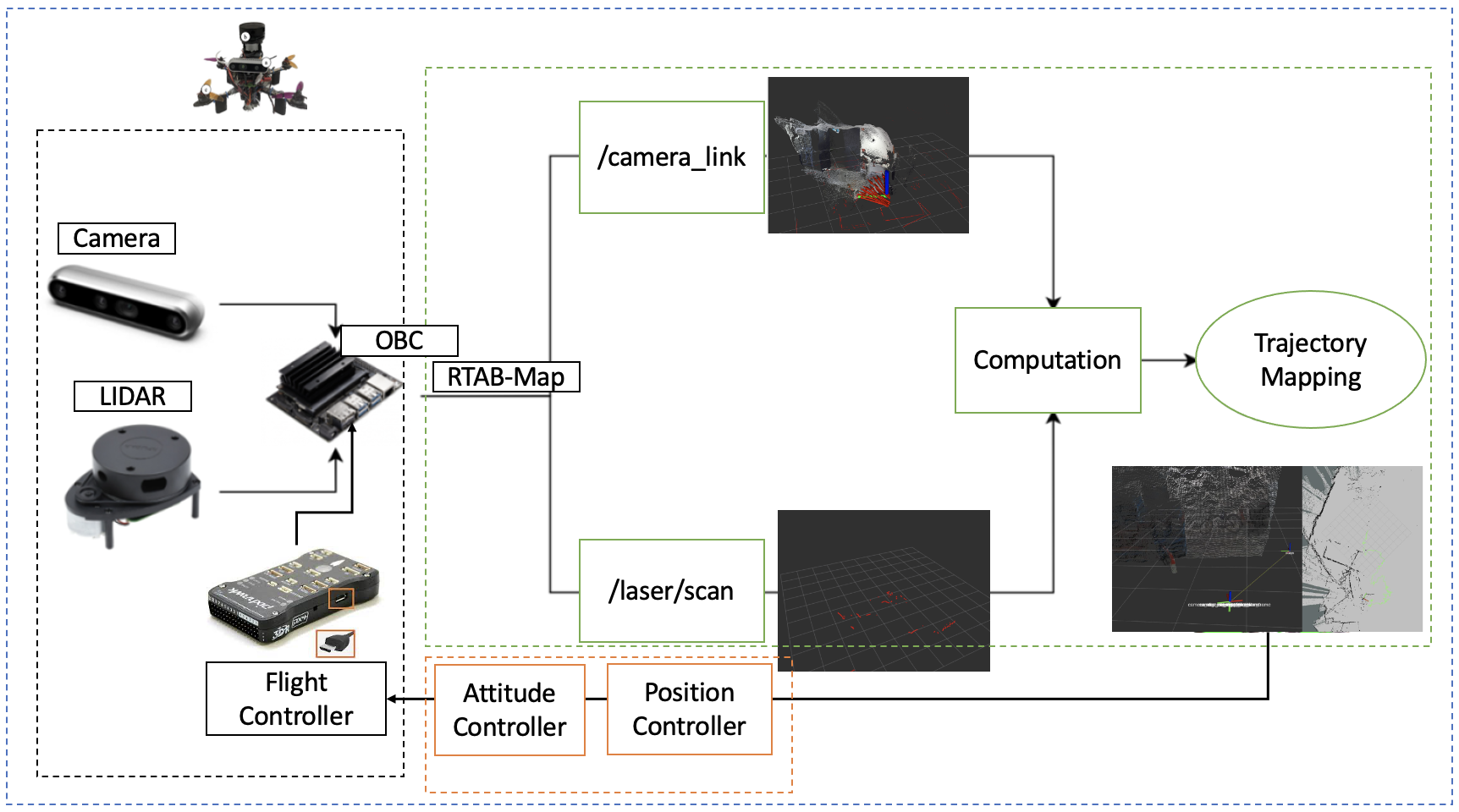}
	\caption{The System Block Diagram of the Autonomous UAV}
\label{block_diag}  
\end{figure}

 We also show the multi-sensor fusion integration of the on board sensors such as LIDAR, IMU, and stereo vision. Fig.~\ref{drone} shows the drone hardware specifications and Fig.~\ref{block_diag} shows the hardware system block diagram of the proposed system. This block diagram is further explained in depth with the deep learning and edge detection in the coming sections. 

In this project, the following system configuration partially adapted from \cite{Seth2023May} is used:

\begin{itemize}
\item   \textbf{UAV Components}: A 295 mm carbon fiber quadrotor frame, 3 bladed 5 inch propellers, 1750 Kv motors, and a 3S 3200 mAh LiPo Battery are used in building the UAV. This gives a balance of stability and control for the UAV with a maximum flight time of 8 minutes with about 1.5 kg total load. A Pixhawk 2.4.8 flight controller with Ardupilot firmware are configured for the UAV.

\item \textbf{On Board Computer (OBC)}: A Jetson Nano board is used as the onboard computing platform. The board runs custom built Ubuntu 20.04 OS, ROS Noetic, and is configured to run Intel and ZED cameras using ROS middleware. The board is powered by a 5V 4A BEC and communicates to the flight controller via serial communication.

\item   \textbf{Sensors}: The ZED 2i stereo camera and the Intel RealSense d455 equipped with dual RGB cameras and an IMU sensors are tested. Both of these are compatible with the software architecture and ROS version making it versatile for applications like SLAM and 3D mapping. A Slamtec RPLidar A1M8 laser scanning is used mainly for confined spaces obstacle detection. Apart from this, the IMU sensors are also used for motion tracking, orientation, and odometry data.

\item \textbf{SLAM System}: RTAB-Map (Real-Time Appearance-Based Mapping) integrates with ROS to execute loop closure detection and construct a global map, by subscribing to odometry, sensor data, and other ROS topics, while publishing the generated 2D or 3D maps and localization data back into the ROS ecosystem. This received data is then converted into attitude, and position controller for the flight controller commands.  

\end{itemize}

\section{Sensor Fusion for Mapping}

\subsection{Sensors}

\textbf{Depth sensing} is a critical component in many robotic and spatial perception applications. The camera is connected to the Jetson Nano through USB 3.0 and uses 1080p Video Mode at 30 fps, detecting distances from 0.3 to 10 meters. The camera provides RGB-D images (\texttt{/rgb\_topic}) and stereo images (\texttt{/depth\_camera\_info\_topic}), which are essential inputs for our sensor fusion block. The (\texttt{roslaunch zed\_rtabmap\_example zed\_rtabmap.launch}) program launches the RTAB-Map program with the appropriate ROS nodes. It transmits data at 1.55 MB/s and 6.18 Hz frequency.

\textbf{Odometry information} is similarly sent using the built-in IMU, Barometer, and Magnetometer sensors in the ZED 2i module. The module provides odometry information (\texttt{zed\_node/odom}) and IMU messages with pose data (\texttt{zed\_node/imu/data}), which are essential inputs for our sensor fusion block. The (\texttt{zed/zed\_node/odom}) ROS topic, transmitting (\texttt{nav\_msgs/Odometry}) messages, operates at a frequency of approximately 7.59 Hz with a bandwidth usage of 5.31 KB/s as shown in Figure~\ref{cam_odom_results_rviz}. The (\texttt{zed/zed\_node/imu/data}) ROS topic, transmitting messages of type (\texttt{sensor\_msgs/Imu}), operates at a frequency of approximately 300.31 Hz with a bandwidth usage of 96.51 KB/s. This data is also fed into the sensor fusion block for synchronization. 

\textbf{LiDAR} information is obtained from the (\texttt{/scan}), type (\texttt{sensor\_msgs/LaserScan}) topic obtained from the RPLidar. The RPLidar laser scanner generates laser scan data representing distance measurements to objects in the UAV's surroundings. This data is published on the (\texttt{/scan}) topic in ROS and can be utilized for various purposes, including obstacle detection, mapping, and localization within the environment. The (\texttt{/scan}) ROS topic, transmitting messages of type (\texttt{sensor\_msgs/LaserScan}), operates at a frequency of approximately 6.4 Hz with a bandwidth usage of 37 kbps.

\subsection{SLAM Visualization in RVIZ}

\begin{figure}[ht]  
    \centering
    \includegraphics[width=1\linewidth]{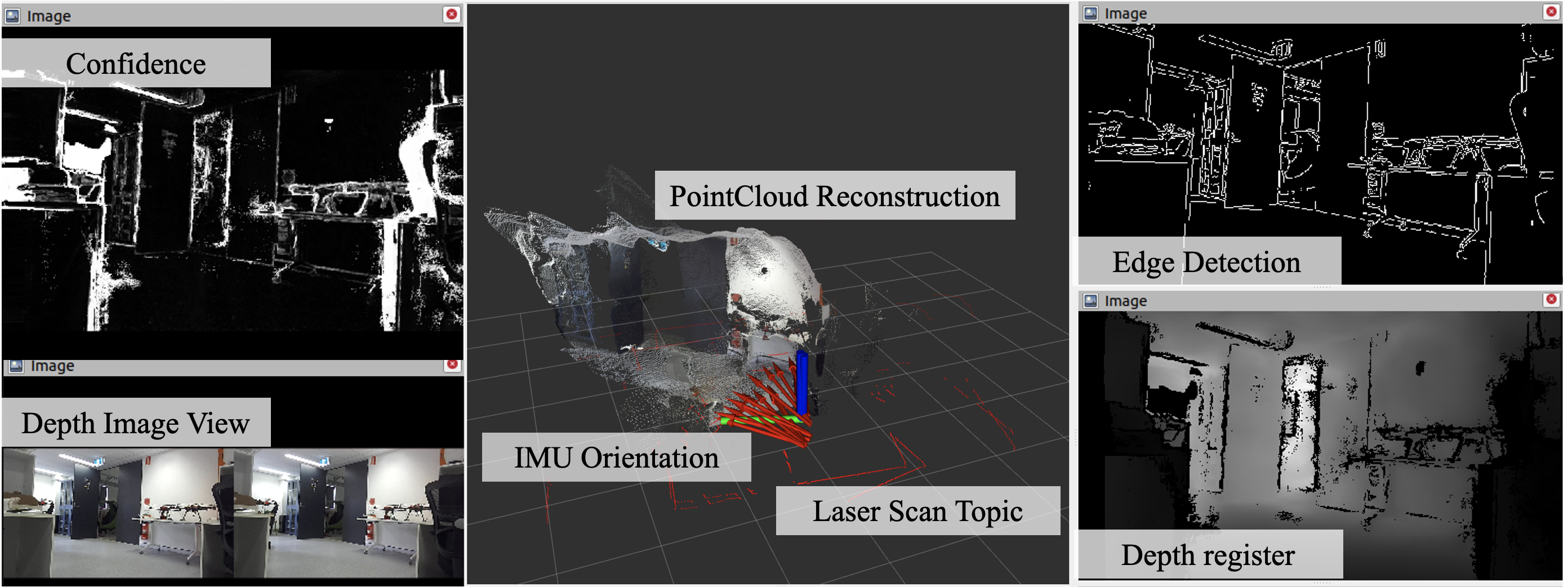}
	\caption{The ROS Visualisation output for SLAM implementation performed by connecting the ground station laptop to the UAV's unique ROS Master IP address}
\label{rviz_results_1}  
\end{figure}

In our \textbf{SLAM} implementation, we employ \textbf{RVIZ}, a ROS visualization tool, to provide real-time insights into the perception and mapping capabilities of our UAV system. This enables us to monitor and analyze various critical aspects of the environment and sensor data as shown in Figure~\ref{rviz_results_1}. 

\textbf{Top Left - Confidence Map:} This section displays the confidence obtained from the ZED 2i camera. The confidence map is crucial for understanding the reliability of the depth information. It is available as an image on the topic \texttt{confidence/confidence\_image}. Note that the Confidence Map is also accessible as a 32-bit floating-point image through the topic \texttt{confidence/confidence\_map}. \textbf{Bottom Left - Depth Image View:} Here, we visualize the depth map obtained from the ZED 2i camera. This depth map is accessible in RVIZ through the topic \texttt{depth/depth\_registered}. It contains 32-bit depth values in meters. RVIZ automatically normalizes the depth map to 8-bit and displays it as a grayscale depth image. \textbf{Top Right - Custom Edge Detection Node:} This section showcases a custom edge detection node, utilizing the Canny edge detection algorithm through the OpenCV ROS node. It subscribes to the ROS topic \texttt{zed\_node/rgb/image\_rect\_color} for image data, allowing us to visualize edge-detection results. \textbf{Bottom Left - Depth Register:} This section is dedicated to visualizing the registered depth information. Registration ensures that depth data aligns with the RGB data, providing accurate spatial information. \textbf{Center of Image - Point Cloud Reconstruction:} At the center of the image, we present a 3D point cloud reconstruction of the environment. This rich representation incorporates information from various sensors, including the ZED 2i camera and RPLidar, as well as the orientation data from the IMU. The point cloud is accessible through the topic (\texttt{zed/zed\_node/point\_cloud/cloud\_reg}).

\par In sensor fusion, loop closure operates by integrating data from multiple sensors to recognize revisited locations or landmarks within an environment \cite{Labbe2018}. This recognition relies on the fusion of sensory information from visual, depth, LiDAR, and IMU data, to establish consistent and accurate maps. When the system detects that the UAV or robotic platform revisits a previously observed location, loop closure algorithms use the fused sensor data to align the new observations with the existing map, thereby correcting mapping errors and ensuring the map remains coherent. 

\par The algorithm~\ref{AL} outlines the steps for indoor localization and navigation of a drone. It starts by initializing odometry and pose estimation, integrating data from an Inertial Measurement Unit (IMU) and a camera. Point cloud data is reconstructed for spatial awareness. When an obstruction or edge is detected, the depth camera's range is set between 0.3 to 10 meters. For objects in close proximity, sensor messages are fetched from a Laser Scan at a frequency of 6.4 Hz. Sensor fusion techniques of loop closure and proximity detection are applied to get map graph data. The drone's velocity and orientation are updated accordingly using MAVROS messages. If the position error is within a certain tolerance, the drone switches to a ($Guided\_No\_GPS$) mode. The final output is a trajectory map for indoor navigation.

\begin{algorithm}[!t]
\begin{algorithmic}[1]
\caption{Drone Indoor Localization and Navigation}
\label{AL}

    \STATE \textbf{Initialize: Odom and Pose Estimation}
    \label{init}
  
    \STATE{Integrate IMU and Camera Feedback for Pose Estimation}
    \STATE{Reconstruct Point Cloud Data from /point\_cloud/cloud\_reg}
   
    \WHILE{Obstruction or Edge Detected}
        \STATE{Set depth\_camera\_info\_topic range to 0.3 to 10 meters}       
        \FOR{Objects in Close Proximity (Camera Range Limited)}
            \STATE{Fetch sensor\_msgs from LaserScan at 6.4 Hz}
            \label{X}   
            \STATE{Perform Sensor Fusion: Loop Closure and Proximity Detection}
            \label{Z}  
            \STATE{Update (velocity\_ned) based on current (drone\_position)}
            \label{Y}            
            \STATE{Estimate Orientation using \(yaw\_rate\_cmd\)}
            \label{B}
            \IF{Position Error within Tolerance}
                \STATE{Switch to Guided\_No\_GPS Mode}
                \label{V}
            \ENDIF 
        \ENDFOR   
    \ENDWHILE
    
    \STATE \textbf{Output}: Generate Trajectory Map for Indoor Environment
    \label{final}
\end{algorithmic}
\end{algorithm}

\section{Experimental Validation}

In our experimental setup, all the computation is done on board the UAV on the Jetson Nano computer. We first verify and test the computational requirements and operation without attaching propellers and rectify any calibration and setup phases necessary for the communication between different sensors, controllers, and actuators present on the UAV. An experimental setup in the real-world is created to test the algorithm and model performance. 

\subsection{System Network - RQT graphs}

\begin{figure}[ht]    
    \centering
    \includegraphics[width=1\linewidth]{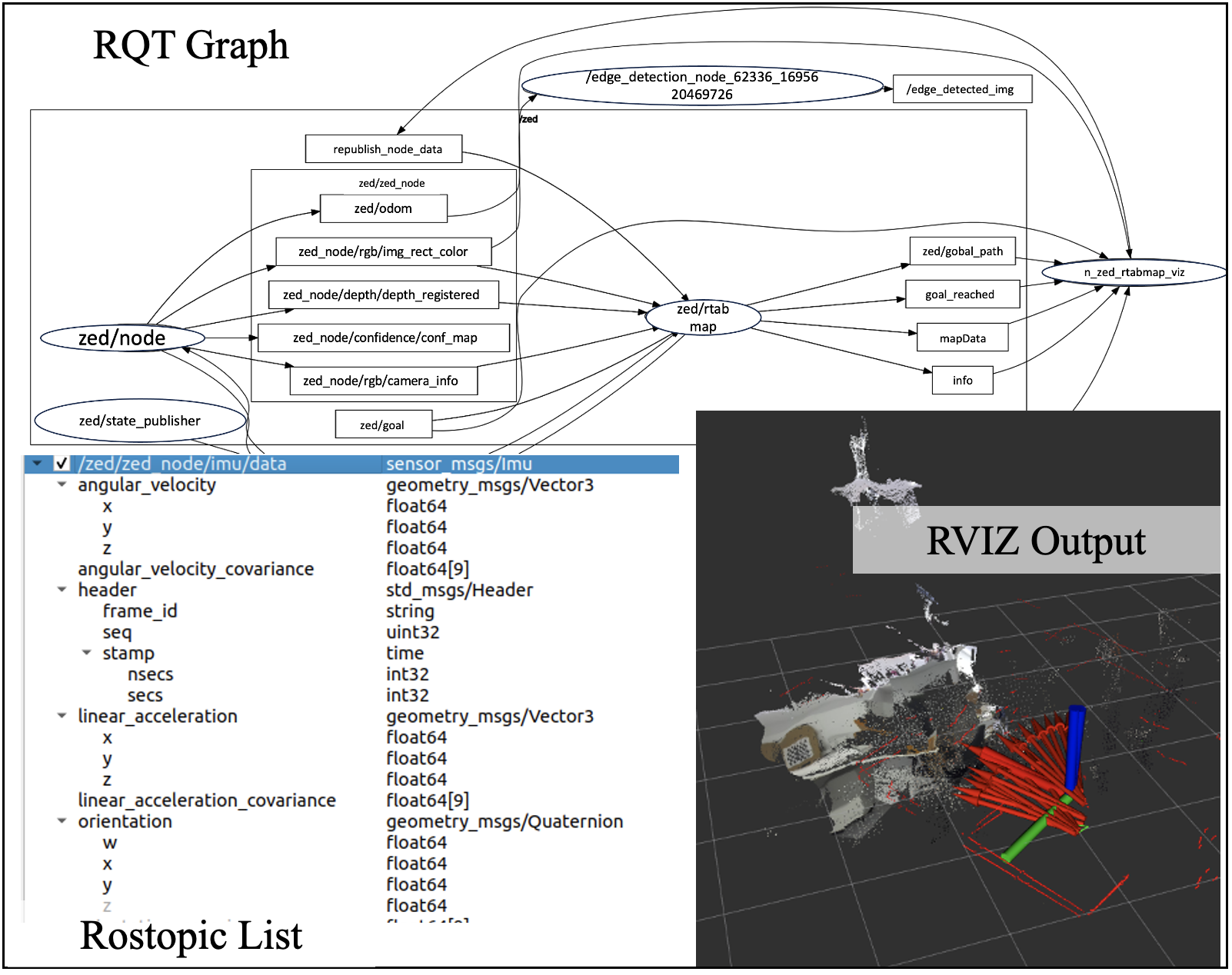}
	\caption{The RQT Graph showing active ROS nodes and topics obtained from the ZED 2i camera along with RVIZ output}
\label{cam_odom_results_rviz}  
\end{figure}

\begin{figure}[htpb!]   
    \centering
    \includegraphics[width=1\linewidth]{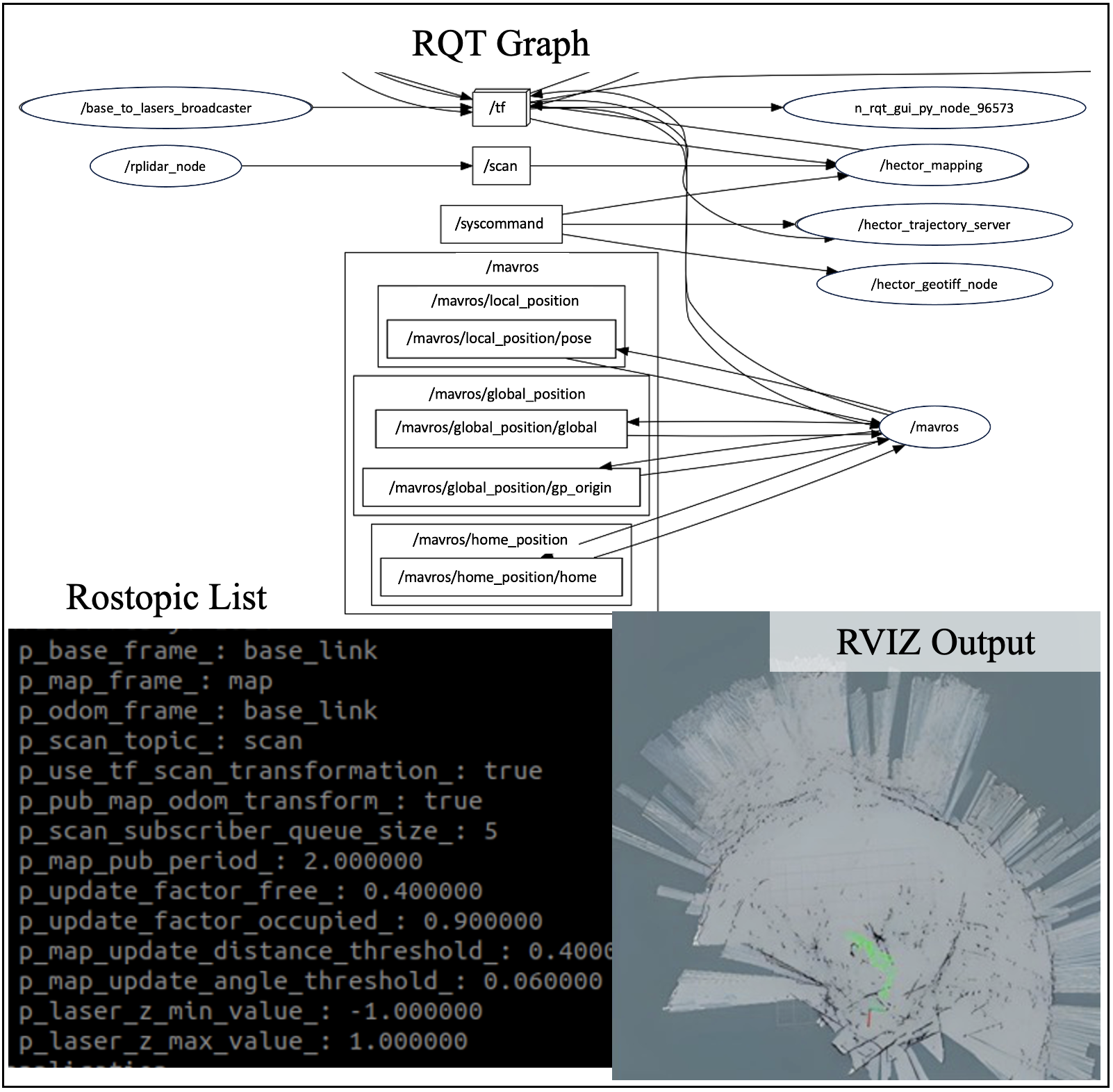}
	\caption{The RQT graph showing the active ROS nodes, ROS topics and TF for the LiDAR, Hector Slam, and MAVROS packages along with the RVIZ output.}
\label{laser_results_2}  
\end{figure}

We use $Rqt\_graph$ to illustrate the communication structure of the components in our autonomous indoor UAV navigation system, showing how different sensors, processing nodes, and the flight controller are connected via ROS topics. The complete $Rqt\_graph$ is divided into two figures for better clarity. Figure~\ref{cam_odom_results_rviz} shows the $Rqt\_graph$ presenting active ROS nodes and topics obtained from the ZED 2i camera along with RVIZ output and Figure~\ref{laser_results_2} shows the $Rqt\_graph$ for mavros and rplidar laser ROS nodes. Explained earlier in section 4, multiple camera ROS nodes are used to create the RTAB-Map graph and visualisation that is used for positioning, orientation and trajectory analysis.

\par Figure~\ref{cam_odom_results_rviz} shows the $Rqt\_graph$ for mavros and rplidar laser ROS nodes. The TF library, denoted by \texttt{(/tf)} allows us to keep track of the relationships between different coordinate frames attached to various components of our UAV in a 3D space. The communication between the mavros node responsible for the flight controller's messages and control outputs and \texttt{(/rplidar)} is shown here. The Rostopic list shows the continuous messages published through the network to maintain the map generation and trajectory shown in green colour in the RVIZ output. Hector SLAM ROS package, when integrated with the RPLidar sensor's $/scan$ data and RViz visualization, enables robots to construct 2D maps of their surroundings while concurrently estimating their own positions, supporting tasks such as autonomous navigation and obstacle avoidance. 

\par
\begin{figure}[htpb!]    
    \centering
    \includegraphics[width=1\linewidth]{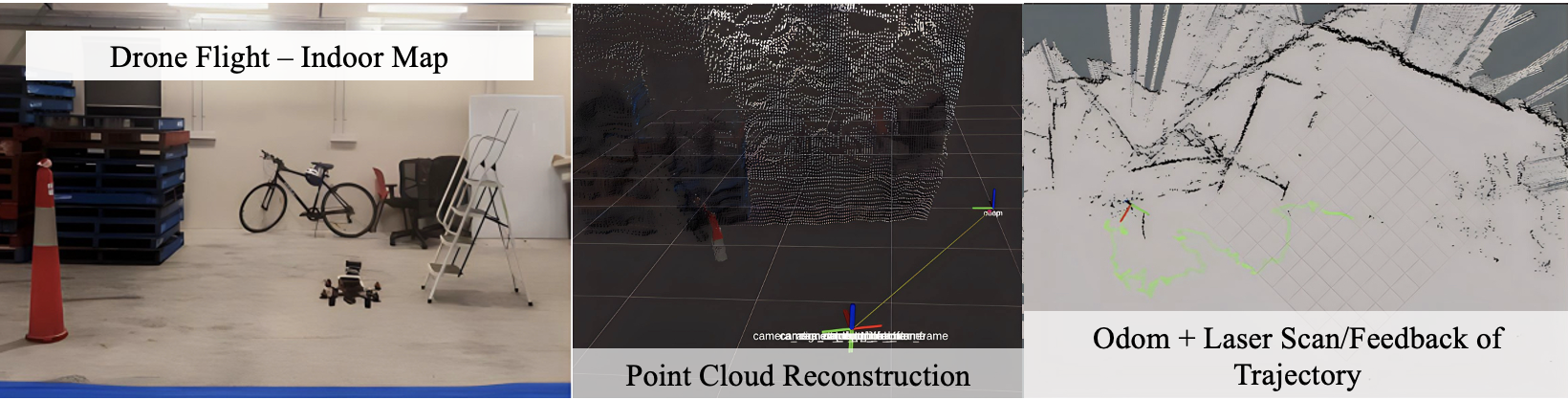}
	\caption{Indoor Autonomous UAV Navigation Flight Tests in a Confined Space setup}
\label{flight_test_rviz_results}  
\end{figure}

 An experimental setup is created using different obstacles placed closely together for the custom UAV to navigate through autonomously. Our real-world experiments were conducted within a warehouse space and customised to roughly 7m × 7m × 5m dimensions as shown in Figure~\ref{flight_test_rviz_results}. The point cloud reconstructed, TF transforms of the links, and coloured confidence map are shown in the middle Figure~\ref{flight_test_rviz_results}. The right most section in Figure~\ref{flight_test_rviz_results} represents the planned trajectory of the UAV in the confined space representing the Odom and Laser Scan/Feedback. The UAV has demonstrated its capability to perform effectively even in close proximity to obstacles. 

\subsection{Distance measurements}

To verify the accuracy and performance of the UAV, a total of 20 trials are performed in the shown indoor space with three different sensor configurations. Table \ref{tab:sensor-comparison} provides a comparison of 3 different sensor configurations for the UAV system navigation. The configurations include "Monocular Camera Only," "Depth Camera + IMU," and "Depth Camera + IMU + LiDAR." The data shows that as the sensor configuration complexity increases, navigation accuracy improves, with the LiDAR-equipped system achieving the highest accuracy up to 0.4 meters. Mapping quality, measured in RMSE, follows a similar trend, with the LiDAR configuration having the lowest RMSE of 0.13 m. However, this improved performance comes at a cost of increased power and energy consumption, where the LiDAR setup consumes 8 watts and 1440 joules, while the monocular camera consumes only 5 watts and 900 joules. The LiDAR configuration is also the heaviest at 1.5 kg, compared to 1 kg for the monocular camera setup. Hence, the choice of sensor configuration needs to be balanced considering a trade-off between accuracy, cost, energy consumption, and weight requirements.

\begin{table}[ht]
\caption{Comparison of 3 different sensor configurations tested on the uav \\
A = monocular camera, B = depth camera + imu, C = depth camera + imu + lidar}
    \resizebox{\columnwidth}{!}
    {\begin{tabular}{|c|c|c|c|}
    \hline
        \textbf{Sensor Configuration} & \textbf{A} & \textbf{B} & \textbf{C} \\ [1ex] \hline
        \textit{Navigation Accuracy (m)} & 0.7 & 0.5 & 0.4 \\ [1ex] \hline
        \textit{Mapping Quality (RMSE)}  & 0.22 & 0.15 & 0.13 \\ [1ex] \hline
        \textit{Exploration Time (min)} & 18 & 12 & 13 \\ [1ex] \hline
        \textit{Average Power Consumption (Watts)}  & 5 & 7 & 8 \\ [1ex] \hline
       \textit{Total Energy Consumption (Joules)}  & 900 & 1260 & 1440 \\ [1ex] \hline
        \textit{Cost (Au\$)}  & \$800  & \$1,300  & \$1,500  \\ [1ex] \hline
       \textit{Weight (kg)}  & 1 kg & 1.2 kg & 1.5 kg \\ [1ex] \hline
    \end{tabular}}
    \label{tab:sensor-comparison}

\end{table}

\begin{figure}[ht]    
    \centering
    \includegraphics[width=1\linewidth]{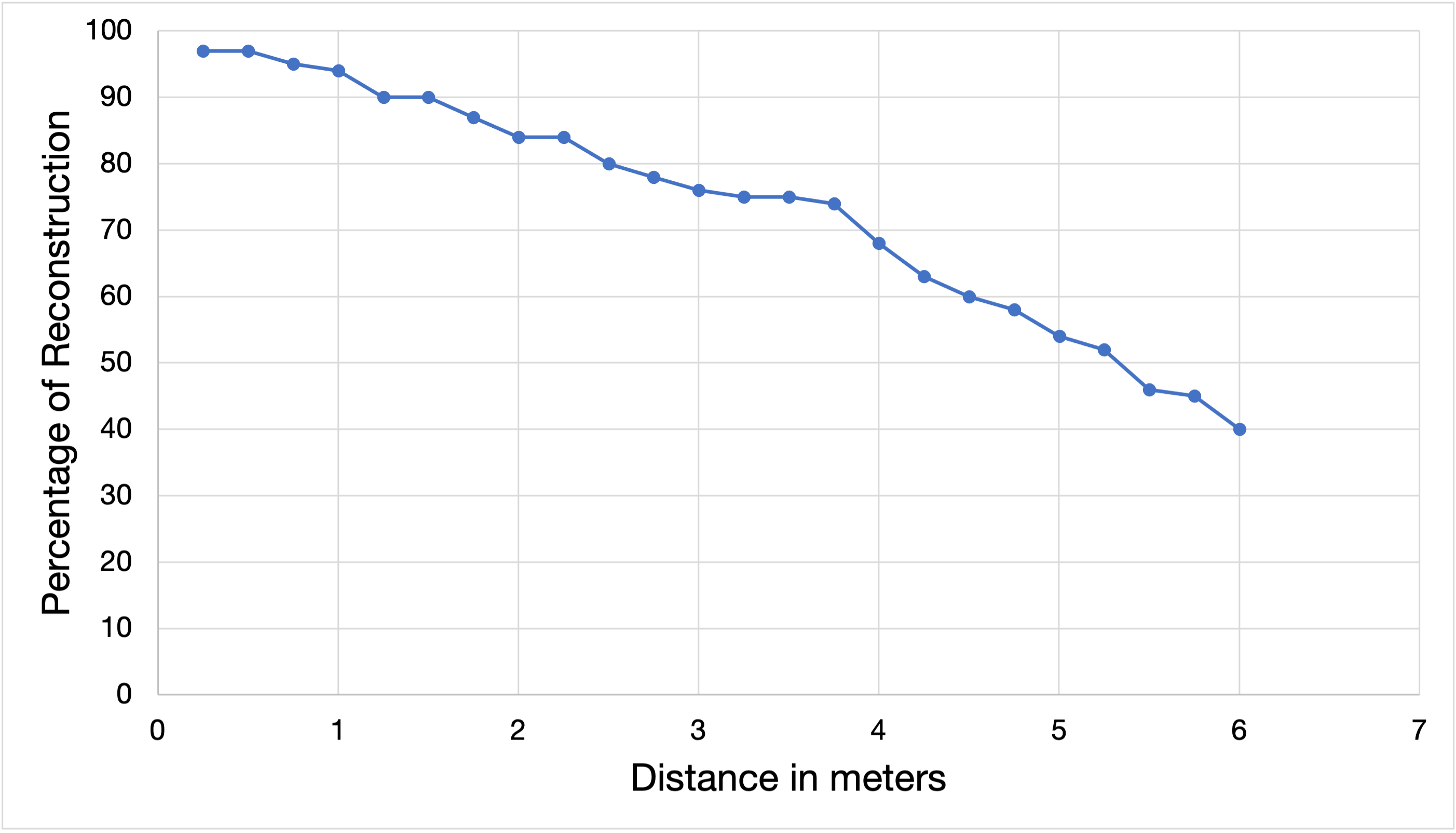}
	\caption{Detection and Reconstruction of the map at different distances}
\label{perc_recon}  
\end{figure}

The provided data in Figure~\ref{perc_recon} illustrates the relationship between distance (measured in meters) and the percentage of successful reconstruction. The results indicate that at shorter distances, the reconstruction performance is consistently high, with 97\% accuracy at 0.25m and 0.5m, gradually declining to 94\% at 1m. Beyond this point, there is a more pronounced decrease in accuracy, with 76\% accuracy at 3m and 68\% at 4m. As the distance increases further, the accuracy continues to decrease, reaching 40\% at 6m. This data suggests that the reconstruction system performs exceptionally well for close-range objects but faces challenges as the distance increases, which proves that the system performs well in close proximity areas.

\subsection{Flight tests}

In this section, we describe the flight tests conducted using our aerial robotic system. The tests were performed in an indoor environment utilizing a custom indoor map as shown earlier. One of the key aspects of these flight tests was the evaluation of the system's orientation control. To assess this, we recorded and analyzed the Roll, Pitch, and Yaw (RPY) IMU readings obtained from the Pixhawk flight controller during flight.

\begin{figure}[ht]    
    \centering
    \includegraphics[width=0.9\linewidth]{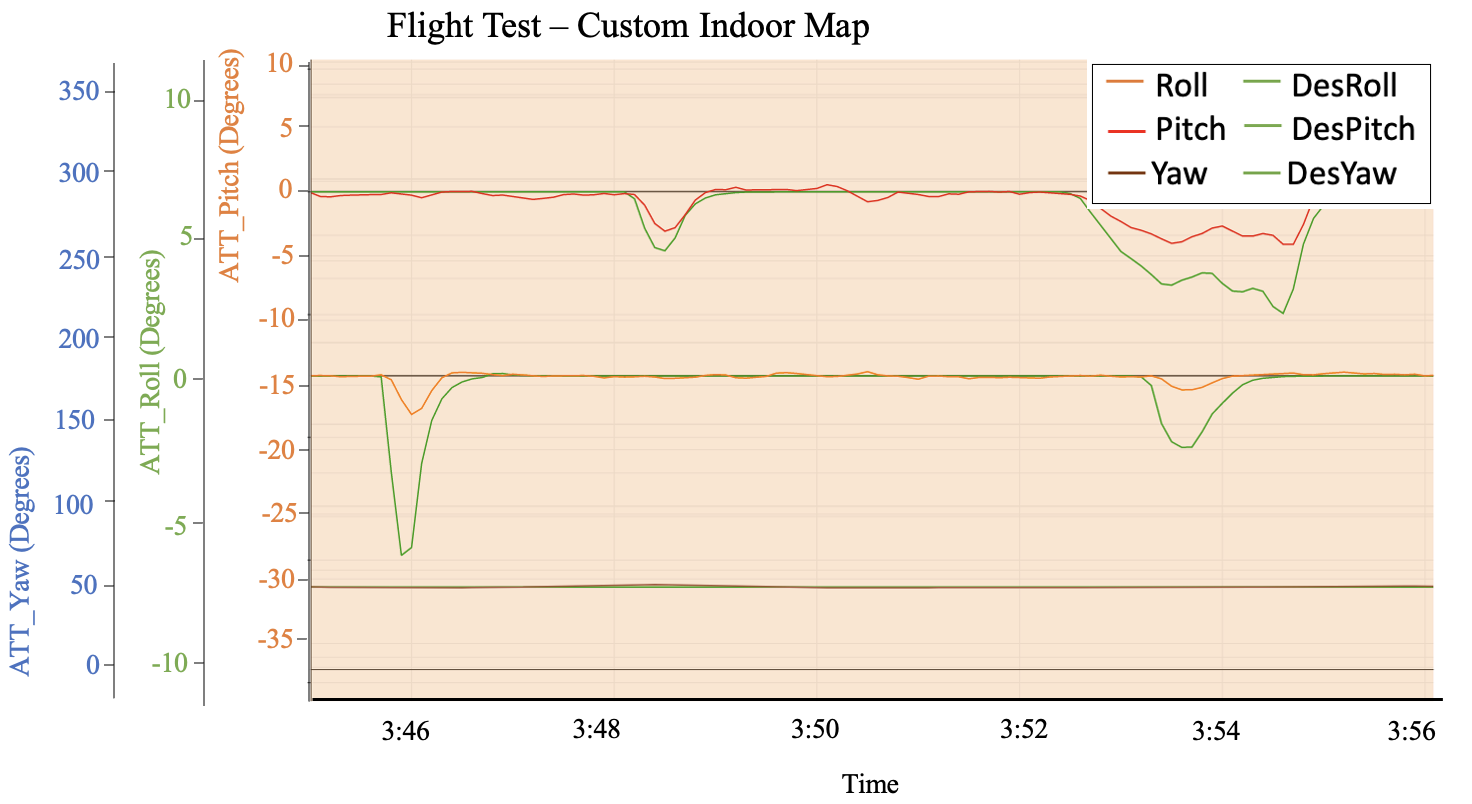}
	\caption{The flight log data of the drone mid-flight highlighting the actual vs. desired Roll, Pitch, and Yaw values for the Autonomous Indoor Navigation Test}
\label{flight_test_results}  
\end{figure}

The flight test results were graphically represented, comparing the desired RPY values with the actual values recorded during flight as shown in Figure~\ref{flight_test_results}. This comparison allowed us to evaluate the system's performance in terms of maintaining the desired orientation. Notably, the error rate observed in these flight tests was found to be impressively low, measuring at just about 0.1\%. This low error rate underscores the precision and reliability of the orientation control system in maintaining the desired flight attitude.

\section{Conclusion}

In conclusion, this paper presents a comprehensive study on the development and validation of a sensor-fusion-based system for autonomous indoor UAV navigation in confined spaces. Through a series of experiments, we have demonstrated the system's robustness and efficacy in real-world scenarios. Our sensor fusion approach, combining depth sensing from the ZED 2i camera, IMU data, and LiDAR measurements, has shown definitive results. The system achieved a navigation accuracy as low as 0.4 meters, a mapping quality with a Root Mean Square Error (RMSE) of only 0.13 m, and a trajectory mapping capability that enables autonomous UAV flight in confined spaces. Importantly, these results are achieved while keeping energy consumption balanced, with the LiDAR-equipped system consuming 1440 joules, and demonstrating the system's reliability and energy efficiency. Furthermore, the flight tests substantiate the system's orientation control, with a negligible error rate of 0.1\%, attesting to its precision in maintaining desired flight attitudes. This work marks a significant step forward in developing autonomous indoor UAV navigation systems, offering practical solutions for applications in search and rescue, facility inspection, and environmental monitoring within GPS-denied indoor environments.

Thus, future research directions may focus on enhancing the system's adaptability and obstacle avoidance capabilities in dynamic indoor settings. In conjunction with this, the research presented in this study establishes a path for the effective use of autonomous UAVs in difficult indoor scenarios, offering major advances in numerous domains where precise, GPS-free navigation is essential.

\bibliographystyle{IEEEtran}
\bibliography{ref_new.bib}

\end{document}